\documentstyle[colacl]{article}
\author{\hspace*{-0.5cm}Masaki Murata \hspace{0.6cm}  Hitoshi Isahara \\
\hspace*{-0.5cm}Communications Research Laboratory\\
\hspace*{-0.5cm}588-2, Iwaoka, Nishi-ku, Kobe, 651-2401, Japan\\
\hspace*{-0.5cm}{\tt \{murata,isahara\}@crl.go.jp}\\
\hspace*{-0.5cm}TEL: +81-78-969-2181, FAX: +81-78-969-2189
\And  \hspace*{2.5cm}Makoto Nagao \\
\hspace*{2.5cm}Kyoto University\\
\hspace*{2.5cm}Sakyo, Kyoto 606-8501, Japan \\
\hspace*{2.5cm}{\tt nagao@pine.kuee.kyoto-u.ac.jp}}
\title{\mbox{Resolution of Indirect Anaphora} \\ in Japanese Sentences\\
Using Examples ``X {\it no} Y (Y of X)''}

\begin{document}
\maketitle
\bibliographystyle{acl}


\begin{abstract}
A noun phrase can indirectly refer to an entity that has already been
mentioned. For example, ``{\it I went into an old house last night. 
{The roof} was leaking badly and ...}'' 
indicates that ``{\it the roof}'' is associated with ``{\it an old house}'',
which was mentioned in the previous sentence. This kind of reference 
 (indirect anaphora) has not been 
studied well in natural language processing, but is important for 
coherence resolution, language understanding, and machine translation. 
In order to analyze indirect anaphora, 
we need a case frame dictionary for nouns 
that contains knowledge of the relationships between two nouns
but no such dictionary presently exists. 
Therefore, we are forced to use examples of ``X {\it no} Y'' (Y of X) 
and a verb case frame dictionary instead. 
We tried estimating indirect anaphora 
using this information 
and obtained a recall rate of 63\% and 
a precision rate of 68\% on test sentences. 
This indicates that 
the information of ``X {\it no} Y'' 
is useful to a certain extent 
when we cannot make use of a noun case frame dictionary. 
We estimated the results that would be given by 
a noun case frame dictionary, 
and obtained recall and precision rates of 
71\% and 82\% respectively. 
Finally, we proposed a way to construct a noun case frame dictionary 
by using examples of ``X {\it no} Y.'' 
\end{abstract}

\section{Introduction}
\label{sec:4c_intro}

A noun phrase can indirectly refer to an entity that has already been
mentioned. 
For example, ``{\it I went into an old house last night. 
{The roof} was leaking badly and ...}'' 
indicates that ``{\it The roof}'' is associated with ``{\it an old house},'' 
which has already been mentioned. This kind of reference 
 (indirect anaphora) has not been thoroughly 
studied in natural language processing, 
but is important for 
coherence resolution, language understanding, and machine translation. 
We propose a method that will resolve the indirect anaphora in Japanese nouns 
by using the relationship between two nouns. 

When we analyze indirect anaphora, 
we need a case frame dictionary for nouns 
that contains information about the relationship between two nouns. 
For instance, in the above example, 
the knowledge that 
``roof'' is a part of a ``house'' is required 
to analyze the indirect anaphora. 
But no such noun case frame dictionary exists at present. 
We considered using the example-based method 
to solve this problem. 
In this case, 
the knowledge that ``roof'' is a part of ``house'' is 
analogous to ``house of roof.'' 
Therefore, 
we use examples of the form ``X of Y'' instead. 
In the above example, 
we use linguistic data such as ``the roof of a house.'' 
In the case of verbal nouns, we do not use 
``X of Y'' but a verb case frame dictionary. 
This is because 
a noun case frame is similar to 
a verb case frame 
and a verb case frame dictionary does exist. 

The next section describes 
a method for resolving indirect anaphora. 

\section{How to Resolve Indirect Anaphora}
\label{how_to}

Anaphors and antecedents in indirect anaphora 
have a certain relationship. 
For example, 
``{\it yane} (roof)'' and ``{\it hurui ie} (old house)'' 
are in an indirect anaphoric relationship  
which is a part-of relationship. 
\begin{equation}
  \begin{minipage}[h]{11.5cm}
    \begin{tabular}[t]{lllll}
      {\it sakuban} & {\it aru}  & {\it hurui} & {\it ie-ni} & {\it itta.}\\
  (last night) & (a certain) & (old) & (house) & (go)\\
\multicolumn{5}{l}{
  (I went into an old house last night.)}
    \end{tabular}

\vspace{0.3cm}

  \begin{tabular}[t]{lll}
    \underline{\it yane-wa} & {\it hidoi}  & {\it amamoride} ... \\
  (roof) & (badly) & (be leaking)\\
\multicolumn{3}{l}{
  (\underline{The roof} was leaking badly and ... )}
    \end{tabular}
  \end{minipage}
  \label{eqn:ie_sakuban}
\end{equation}
When we analyze indirect anaphora, 
we need a dictionary containing 
information about relationships between anaphors and their antecedents. 

\begin{table*}[t]
\small
  \leavevmode
  \begin{center}
    \caption{Relationships between anaphors and their antecedents}
    \label{tab:noun_case_frame}
\begin{tabular}{|l@{ }|l@{ }|l@{ }|}\hline
Anaphor      & Possible antecedents & Relationship \\\hline
{\it kazoku} (family)  & {\it hito} (human)           & belong\\\hline
{\it kokumin} (nation) & {\it kuni} (country)         & belong\\\hline
{\it genshu} (the head of state)   & {\it kuni} (country) & belong\\\hline
{\it yane} (roof)      & {\it tatemono} (building)    & part of\\\hline
{\it mokei} (model)    & {\it seisanbutsu} (product) & object\\
                & [ex. {\it hikouki} (air plain), {\it hune} (ship)] & \\\hline
{\it gyouji} (event)   & {\it soshiki} (organization) & agent\\\hline
{\it jinkaku} (personality)  & {\it hito} (human)    & possessive\\\hline
{\it kyouiku} (education)    & {\it hito} (human)    & agent\\
             & {\it hito} (human)             & recipient\\
             & {\it nouryoku} (ability)     & object\\
             & [ex. {\it suugaku} (mathematics)]  & \\\hline
{\it kenkyuu} (research) & {\it hito} (human), {\it soshiki} (organization) & agent\\
             & {\it gakumon bun'ya} (field of study)          & object\\\hline
{\it kaiseki} (analysis) & {\it hito} (human), {\it kikai} (machine) & agent\\
             & {\it de-ta} (data)          & object\\\hline
\end{tabular}
\end{center}
\end{table*}

We show examples of the relationships between anaphors and antecedents 
in Table \ref{tab:noun_case_frame}. 
The form of Table \ref{tab:noun_case_frame} is 
similar to the form of a verb case frame dictionary. 
We would call a dictionary containing 
the relationships between two nouns {\sl a noun case frame dictionary} 
but no noun case frame dictionary has yet been created. 
Therefore, we substitute it with examples of ``X {\it no} Y'' (Y of X) 
and with a verb case frame dictionary. 
``X {\it no} Y'' is a Japanese expression. 
It means ``Y of X,'' ``Y in X,'' ``Y for X,'' etc. 

\begin{table*}[t]
  \begin{center}
  \caption{Case frame of verb ``{\it kaiseki-suru} (analyze)''}
  \label{tab:kuitigau_frame}
    \leavevmode
\begin{tabular}[h]{|l|l|l|}\hline
Surface case      & Semantic constraint & Examples\\\hline
{\it ga}-case (subject) & human & {\it seito} (student), {\it kare} (he)\\
{\it wo}-case (object)  & abstract, product & {\it atai} (value), {\it de-ta} (data)\\\hline
\end{tabular}\\
  \end{center}
\end{table*}

\begin{table*}[t]
  \caption{The weight as topic}
  \label{fig:shudai_omomi}
  \begin{center}
    \leavevmode
\begin{tabular}[c]{|p{8.5cm}|l|r|}\hline
\multicolumn{1}{|c|}{Surface expression}  & \multicolumn{1}{|c|}{Example}   & Weight \\\hline\hline
Pronoun/zero-pronoun {\it ga/wa}&  [\underline{John} {\it ga} ({\sf subject})] {\it shita} (done). &21 \\\hline
Noun {\it wa/niwa}        &  \underline{John} {\it wa} ({\sf subject}) {\it shita} (done). &20 \\\hline
\end{tabular}
  \end{center}
\end{table*}

We resolve the indirect anaphora using the following steps: 
\begin{enumerate}
\item 
\label{enum:youso_kenshutu}
We detect some elements which could be analyzed 
by indirect anaphora resolution 
using ``X {\it no} Y'' 
and a verb case frame dictionary. 
When a noun was a verbal noun, 
we use a verb case frame dictionary. 
Otherwise, 
we use examples of ``X {\it no} Y.'' 

For example, 
in the following example sentences 
{\it kaiseki} (analysis) is a verbal noun, and 
we use a case frame of a verb {\it kaiseki-suru} (analyze) 
for the indirect anaphora resolution of {\it kaiseki} (analysis). 
The case frame is shown in Table \ref{tab:kuitigau_frame}. 
In this table there are two case components, 
the {\it ga}-case (subject) and the {\it wo}-case (object). 
These two case components are 
elements which will be analyzed in indirect anaphora resolution. 

\begin{equation}
  \begin{minipage}[h]{6cm}
    \small
\hspace*{-0.8cm}
    \begin{tabular}[t]{l@{ }l@{ }l}
      {\it denkishingou-no} & {\it riyouni-ni} & {\it yotte} \\
  (electronic detectors)  & (use)   & (by) \\
\multicolumn{3}{l}{
  (By using electronic detectors. )}
    \end{tabular}

\vspace{0.3cm}

\hspace*{-0.8cm}
\begin{tabular}[t]{l@{ }l@{ }l}
  {\it Butsurigakusha-wa} & {\it tairyou-no} & {\it deeta-wo}\\
  (physicist)  & (a large amount)   & (data) \\
\end{tabular}

\hspace*{-0.8cm}
\begin{tabular}[t]{l}
  {\it shuushuudekiru-youni-natta.}\\
  (collect)\\
  \multicolumn{1}{p{6cm}}{
    (physicists had been able to collect large amounts of data. )}
\end{tabular}
\end{minipage}
  \label{eqn:kuichigai}
\end{equation}

\hspace*{-0.5cm}
    \begin{tabular}[t]{l@{ }l@{ }l@{ }l@{ }l}
\footnotesize  {\it sokode} & \footnotesize {\it subayai} & \footnotesize {\it \underline{kaiseki}-no} & \footnotesize {\it houhou-ga} & \footnotesize {\it hitsuyouni-natta.}\\
  (then)  & (quick) & (analysis) & (method) & (require)\\
\multicolumn{5}{l}{
  (Then, they required a method of quick \underline{analysis}.)}
    \end{tabular}

\item 
\label{enum:kouho_age}
We take possible antecedents 
from topics or foci in the previous sentences. 
We assign them a certain weight based on 
the plausibility that they are antecedents. 
The topics/foci and their weights are 
defined in Table \ref{fig:shudai_omomi} and Table \ref{fig:shouten_omomi}. 

\begin{table*}[t]
  \caption{The weight as focus}
  \label{fig:shouten_omomi}
  \begin{center}
    \leavevmode
\begin{tabular}[c]{|p{8.5cm}|l|r|}\hline
\multicolumn{1}{|p{8.5cm}|}{Surface expression (Not including ``{\it wa}'')}  & \multicolumn{1}{|c|}{Example}   & Weight \\\hline\hline
Pronoun/zero-pronoun {\it wo} ({\sf object})/{\it ni} (to) /{\it kara} (from) & [\underline{John} {\it ni} (to)] {\it shita} (done). & 16 \\\hline
Noun {\it ga} ({\sf subject})/{\it mo}/{\it da}/{\it nara}/{\it koso} & \underline{John} {\it ga} ({\sf subject}) {\it shita} (done).  & 15 \\\hline
Noun {\it wo} ({\sf object})/{\it ni}/, /.       & \underline{John} {\it ni} ({\sf object}) {\it shita} (done).   & 14 \\\hline
Noun {\it he} (to)/{\it de} (in)/{\it kara} (from)/{\it yori}    & \underline{{\it gakkou} (school)} {\it he} (to) {\it iku} (go).  & 13 \\[0.1cm]\hline
\end{tabular}
  \end{center}
\end{table*}

\begin{table*}[t]
  \caption{Case frame of verb ``{\it mukau}'' (go to)}
  \label{tab:mukau_frame}
  \begin{center}
    \leavevmode
\begin{tabular}[h]{|l|l|l|}\hline
Surface case      & Semantic constraint & Examples\\\hline
{\it ga}-case ({\sf subject})  & concrete         & {\it kare} (he), {\it hune} (ship)\\
{\it ni}-case ({\sf object})   & place            & {\it kouen} (park), {\it minato} (port)\\\hline
\end{tabular}\\
  \end{center}
\end{table*}

For example, in the case of 
``{\it I went into an old house last night. {The roof} was leaking badly and ...},''  
``an old house'' becomes a candidate of the desired antecedent. 
In the case of ``analysis'' in example sentence \ref{eqn:kuichigai}, 
``electronic detectors,'' ``physicists,'' and ``large amounts of data'' 
become candidates of the two desired antecedents of ``analysis.'' 
In Table \ref{fig:shudai_omomi} and Table \ref{fig:shouten_omomi} 
such candidates are given certain weights which indicate preference. 

\item 
We determine the antecedent 
by combining 
the weight of topics and foci 
mentioned in step \ref{enum:kouho_age}, 
the weight of 
semantic similarity in ``X {\it no} Y'' or in a verb case frame dictionary, 
and the weight of the distance 
between an anaphor and its possible antecedent. 

For example, when we want to clarify the antecedent of {\it yane} (roof) 
in example sentence \ref{eqn:ie_sakuban}, 
we gather examples of ``{\sf Noun X} {\it no yane} (roof)'' 
 (roof of {\sf Noun X}), 
and select a possible noun 
which is semantically similar to {\sf Noun X} as its antecedent. 
In example sentence \ref{eqn:kuichigai}, 
when we want to have an antecedent of {\it kaiseki} (analysis) 
we select as its antecedent a possible noun 
which satisfies the semantic constraint in the case frame of 
{\it kuichigau} (differ) in Table \ref{tab:kuitigau_frame} 
or is semantically similar to 
examples of components in the case frame. 
In the {\it ga}-case ({\sf subject}), 
of three candidates, ``electronic detectors,'' ``physicists,'' and ``large amounts of data,'' 
only ``physicists'' satisfies the semantic constraint, 
{\sf human}, in the case frame of the verb {\it kaiseki-suru} in Table \ref{tab:kuitigau_frame}. 
So ``physicists'' is selected as the desired antecedent of the {\it ga}-case. 
In the {\it wo}-case ({\sf object}), 
two phrases, ``electronic detectors'' and ``large amounts of data'' satisfy 
the semantic constraints, {\sf abstract} and {\sf product}. 
By using the examples ``value'' and ``data'' in the case frame, 
the phrase ``large amounts of data,'' which 
is semantically similar to ``data'' in the examples of the case frame, 
is selected as the desired antecedent of the {\it wo}-case. 
\end{enumerate}

We think that 
errors made by the substitution of a verb case frame for a noun case frame are rare, 
but 
many errors occur 
when we substitute ``X {\it no} Y'' for a noun case frame. 
This is because 
``X {\it no} Y'' (Y of X) has many semantic relationships, 
in particular a feature relationship (ex. ``a man of ability''), 
which cannot be an indirect anaphoric relationship. 
To reduce the errors, we use the following procedures. 
\begin{enumerate}
\item 
We do not use an example of the form ``{\sf Noun X} {\it no} {\sf Noun Y}'' (Y of X), 
when noun X is an adjective noun [ex. HONTOU (reality)], 
a numeral, or a temporal noun. 
For example, we do not use {\it hontou} (reality) {\it no} (of) {\it hannin} (criminal) (a real criminal).  
\item 
We do not use an example of the form ``{\sf Noun X} {\it no} {\sf Noun Y}'' (Y of X), 
when noun Y is a noun 
that cannot be an anaphor of an indirect anaphora. 
For example, we do not use 
``{\sf Noun X} {\it no} {\it tsuru} (crane),'' or ``{\sf Noun X} {\it no} {\it ningen} (human being).'' 
\end{enumerate}
We cannot completely avoid errors 
by introducing the above procedure, 
but we can reduce them to a certain extent. 

Nouns such as {\it ichibu} (part), 
{\it tonari} (neighbor) and {\it betsu} (other) 
need further consideration. 
When such a noun is a case component of a verb, 
we use information on the semantic constraints of the verb. 
We use a verb case frame dictionary as shown in Table \ref{tab:mukau_frame}. 
\begin{equation}
  \begin{minipage}[h]{11.5cm}
    \begin{tabular}[t]{l@{ }l@{ }l@{ }l}
      {\it takusan-no} & {\it kuruma-ga}  & {\it kouen-ni} & {\it tomatte-ita.}\\
      (many) & (car)  & (in the park) & (there were)\\
\multicolumn{4}{l}{
  (There were many cars in the park.)}\\
\end{tabular}

\vspace{0.3cm}

    \begin{tabular}[t]{lll}
      {\it \underline{ichibu}-wa} & {\it kita-ni}  & {\it mukatta}\\
      {[A part (of them)]} & (to the north)  & (went) \\
\multicolumn{3}{l}{
  (A \underline{part} of them went to the north.)}\\
    \end{tabular}
  \end{minipage}
\label{eqn:kuruma_itibu}
\end{equation}
In this example, since {\it ichibu} (part) is a {\it ga}-case (subject) 
of a verb {\it mukau} (go),  
we consult the {\it ga}-case (subject) of the case frame of {\it mukau} (go).  
Some noun phrases which can also be used in the case component 
are written in the {\it ga}-case (subject) of the case frame. 
In this case, 
{\it kare} (he) and {\it hune} (ship) are written as examples of things
which can be used in the case component. 
This indicates that 
the antecedent is semantically similar to 
{\it kare} (he) and {\it hune} (ship). 
Since {\it takusan no kuruma} (many cars) is semantically similar to 
{\it hune} (ship) in the meaning of vehicles, 
it is judged to be the proper antecedent. 

When such a noun as {\it tonari} (neighbor or next) modifies a noun X 
as {\it tonari no} X, 
we consider the antecedent to be a noun 
which is similar to noun X in meaning. 
\begin{equation}
\hspace*{-.5cm}
  \begin{minipage}[h]{7.5cm}
\small
\footnotesize 
    \begin{tabular}[t]{l@{ }l@{ }l@{ }l}
      {\it ojiisan-wa} & {\it ooyorokobi-wo-shite} & {\it ie-ni} & {\it kaerimashita.}\\
      (the old man) & (in great joy)  & (house) & (returned)\\
\multicolumn{4}{l}{
  [The old man returned home (house) in great joy,]}\\
\end{tabular}

\vspace{0.3cm}

\begin{tabular}[t]{l@{ }l@{ }l@{ }l}
  {\it okotta} & {\it koto-wo} & {\it hitobito-ni} & {\it hanashimashita}\\
 (happened to him) & (all things) & (everybody) & (told)\\
\multicolumn{4}{l}{
  (and told everybody all that had happened to him.)}\\
    \end{tabular}

\vspace{0.3cm}

    \begin{tabular}[t]{l@{ }l@{ }l@{ }l@{ }l@{ }l@{ }l}
    {\it \underline{tonari}-no} & {\it ie-ni} & {\it ojiisan-ga} & {\it mouhitori} & {\it sunde-orimashita.}\\
      (next) & (house) & (old man) & (another) & (live)\\
\multicolumn{7}{l}{
  (There lived in the \underline{next} house another old man.)}\\
    \end{tabular}
  \end{minipage}
\label{eqn:tonari_ie}
\end{equation}
For example, when {\it tonari} (neighbor or next) 
modifies {\it ie} (house),  
we judge that the antecedent of {\it tonari} (neighbor or next) is 
{\it ie} (house) in the first sentence. 

\section{Anaphora Resolution System}

\subsection{Procedure}
\label{wakugumi}

Before starting the anaphora resolution process, 
the syntactic structure analyzer transforms 
sentences into dependency structures \cite{kuro}. 
Antecedents are 
determined by heuristic rules for each noun from left to right in the sentences. 
Using these rules, our system gives possible antecedents points, and it determines that 
the possible antecedent having the maximum total score is the desired antecedent. 
This is because a several types of information are combined in anaphora resolution. 
An increase in the points of a possible antecedent corresponds to 
an increase of the plausibility of the possible antecedent. 

The heuristic rules are given in the following form:

\begin{center}
    \begin{minipage}[c]{10cm}
      \hspace*{0.7cm}{\sl Condition} $\Rightarrow$ \{ {\sl Proposal, Proposal,} ... \}\\
      \hspace*{0.7cm}{\sl Proposal} := ( {\sl Possible-Antecedent,} \, {\sl Point} )
    \end{minipage}
\end{center}

\noindent
Surface expressions, semantic constraints, 
referential properties, for example, are written as conditions in the {\sl Condition} part. 
A possible antecedent is written in the {\sl Possible-Antecedent} part. 
{\sl Point} refers to the plausibility of the possible antecedent. 

To implement the method mentioned in Section 2, 
we use the weights $W$ of topics and foci, 
the distance $D$, the definiteness $P$, and the semantic similarity $S$ 
 (in R4 of Section \ref{sec:ref_pro}) to determine points. 
The weights $W$ of topics and foci are given 
in Table~\ref{fig:shudai_omomi} and Table~\ref{fig:shouten_omomi} 
respectively in Section \ref{how_to}, 
and represent the preferability of the desired antecedent. 
In this work, a topic is defined as 
a theme which is described, and 
a focus is defined as 
a word which is stressed by the speaker (or the writer). 
But we cannot detect topics and foci correctly. 
Therefore we approximated them 
as shown in Table~\ref{fig:shudai_omomi} and Table~\ref{fig:shouten_omomi}. 
The distance $D$ is the number of the topics (foci) 
between the anaphor and a possible antecedent 
which is a topic (focus). 
The value $P$ is given by the score of the definiteness 
in referential property analysis \cite{match}. 
This is because it is easier for a definite noun phrase 
to have an antecedent than for an indefinite noun phrase to have one. 
The value $S$ is the semantic similarity 
between a possible antecedent and {\sf Noun X} of ``{\sf Noun X} {\it no} {\sf Noun Y}.'' 
Semantic similarity is shown by level 
in {\it Bunrui Goi Hyou} \cite{bgh}. 


\subsection{Heuristics for determining antecedents}
\label{sec:ref_pro}

We wrote 15 heuristic rules for noun phrase anaphora resolution. 
Some of the rules are given below: 

\begin{enumerate}
\item[R1]  
  When the referential property
  of a noun phrase (an anaphor) is definite, 
  and the same noun phrase A has already appeared, 
  $\Rightarrow$
  \\ \{ (the noun phrase A, \,$30$)\}

  A referential property is estimated by this method \cite{match}.
  This is a rule for direct anaphora. 

\item[R2]  
  When the referential property of a noun phrase is generic, $\Rightarrow$\\
  \{ (generic, \,$10$)\} 

\item[R3] 
  When the referential property of a noun phrase is indefinite, $\Rightarrow$\\
  \{ (indefinite, \,$10$)\}

\item[R4]  
  When a noun phrase Y is not a verbal noun, $\Rightarrow$\\
  \{
  (A topic which has the weight $W$ and the distance $D$, \, $W-D+P+S$),\\ 
  (A focus which has the weight $W$ and the distance $D$, \, $W-D+P+S$),\\ 
  (A subject in a subordinate clause or a main clause of the clause, \, $23+P+S$)\\
  where the values $W$, $D$, $P$, and $S$ are 
  as they were defined in Section \ref{wakugumi}. 

\begin{figure*}[t]

  \begin{center}
\fbox{
\begin{minipage}[h]{15cm}

    \begin{tabular}[t]{lll}
  {\it kono dorudaka-wa}     & {\it kyoutyou-wo} & {\it gikushaku saseteiru.}\\
  (The dollar's surge) & (cooperation)       & (is straining)\\
\multicolumn{3}{l}{
  (The dollar's surge is straining the cooperation. )}
    \end{tabular}
    \begin{tabular}[t]{lllll}
  {\it jikokutuuka-wo}        & {\it mamorouto}      & {\it nisidoku-ga}   & {\it \underline{kouteibuai}-wo} & {\it hikiageta.}\\
  (own currency)       & (to protect)   & (West Germany)& (official rate)& (raised)  \\
\multicolumn{5}{l}{
  (West Germany raised (its) \underline{official rate} to protect the mark. )}
    \end{tabular}

\vspace{0.5cm}

\begin{tabular}[h]{|l|l@{ }|r@{ }|r@{ }|r@{ }|r@{ }|r@{ }|}\hline
\multicolumn{2}{|l|}{}        & Indefinite  & {\it nisidoku}    & {\it jikokutuuka} &  {\it kyoutyou} & {\it dorudaka} \\\hline
\multicolumn{2}{|l|}{}        &             & West Germany& own currency    & cooperation      & dollar's surge \\\hline
\multicolumn{2}{|l|}{R3}      &   10        &             &           &            &           \\\hline
\multicolumn{2}{|l|}{R4}      &             &   25        &  $-23$    &  $-24$     &  $-17$    \\\hline
       &Subject             &             &   23        &           &            &           \\
      &Topic Focus $ (W)$      &             &             &    14     &    14      &    20     \\
     &Distance $ (D)$          &             &             &   $-2$    &   $-3$     &   $-2$    \\
           &Definiteness $ (P)$&             &  $-5$       &   $-5$    &   $-5$     &   $-5$    \\
            &Similarity $ (S)$ &             &    7        &  $-30$    &  $-30$     &   $-30$   \\\hline
\multicolumn{2}{|l|}{Total Score} &   10    &   25        &  $-23$    &  $-24$     &   $-17$    \\\hline
\end{tabular}

\vspace{0.5cm}

Examples of ``noun X {\it no kouteibuai} (official rate)''

\vspace{0.5cm}

\hspace{0.5cm}
``{\it nihon} (Japan) {\it no kouteibuai} (official rate)'',

\hspace{0.5cm}
``{\it beikoku} (USA) {\it no kouteibuai} (official rate)''

\caption{Example of indirect anaphora resolution}
\label{tab:dousarei}
\end{minipage}
}
  \end{center}
\end{figure*}

\item[R5]  
  \label{enum:sahenmeishi}
  When a noun phrase is a verbal noun, $\Rightarrow$\\
  \{
  (A topic which satisfies the semantic constraint in a verb case frame and has the weight $W$ and the distance $D$, \, $W-D+P+S$),\\ 
  (A focus which satisfies the semantic constraint and has the weight $W$ and the distance $D$, \, $W-D+P+S$),\\ 
  (A subject in a subordinate clause or a main clause of the clause, \, $23+P+S$)\\
 
\item[R6]   
  When a noun phrase is a noun such as {\it ichibu}, {\it tonari},
  and it modifies a noun X, $\Rightarrow$\\
  \{ (the same noun as the noun X,  \, $30$)\}

\end{enumerate}

\subsection{Example of analysis}

An example of the resolution of an indirect anaphora is shown 
in Figure \ref{tab:dousarei}. 
Figure \ref{tab:dousarei} shows that 
the noun {\it koutei buai} (official rate) is analyzed well. 
This is explained as follows: 

The system estimated the referential property of 
{\it koutei buai} (official rate) to be indefinite in the method \cite{match}. 
Following rule R3 (ection \ref{sec:ref_pro})  
the system took a candidate ``Indefinite,'' 
which means that 
the candidate is an indefinite noun phrase that 
does not have an indirect anaphoric referent. 
Following R4 (Section \ref{sec:ref_pro}) 
the system took four possible antecedents, 
{\it nisidoku} (West Germany), {\it jikokutuuka (own currency)}, {\it kyoutyou} (cooperation), 
{\it dorudaka} (dollar's surge). 
The possible antecedents were given points based on 
the weight of topics and foci, the distance from the anaphor, and so on. 
The system properly judged that 
{\it nisidoku} (West Germany), which had the best score, 
was the desired antecedent.

\section{Experiment and Discussion}

Before the antecedents in indirect anaphora were determined, 
sentences were transformed into a case structure 
by the case analyzer \cite{kuro}. 
The errors made by the analyzer were corrected by hand. 
We used the IPAL dictionary \cite{ipal} as a verb case frame dictionary. 
We used the Japanese Co-occurrence Dictionary \cite{edr_kyouki_1.0} 
as a source of examples for ``X {\it no} Y.'' 

\begin{table*}[t]
\fbox{
\begin{minipage}[h]{16cm}
    \caption{Results}
    \label{tab:sougoukekka}
\vspace*{0.5cm}
  \begin{center}
\begin{tabular}[c]{|r@{}c@{ }|r@{}c@{ }|r@{}c@{ }|r@{}c@{ }|r@{}c@{ }|r@{}c@{ }|}\hline
        \multicolumn{4}{|c|}{Non-verbal noun} &\multicolumn{4}{c|}{Verbal noun}  &\multicolumn{4}{c|}{Total}\\\cline{1-12}
        \multicolumn{2}{|c|}{Recall}
        &\multicolumn{2}{c|}{Precision}
        &\multicolumn{2}{c|}{Recall}
        &\multicolumn{2}{c|}{Precision}
        &\multicolumn{2}{c|}{Recall}
        &\multicolumn{2}{c|}{Precision}\\\hline
\multicolumn{12}{|c|}{Experiment made when the system does not use any semantic information}\\\hline
 85\% & (56/66) & 67\% & (56/83) & 40\% & (14/35) & 44\% & (14/32) & 69\% & (70/101)& 61\% & (70/115)\\\hline
 53\% & (20/38) & 50\% & (20/40) & 47\% & (15/32) & 42\% & (15/36) & 50\% & (35/70) & 46\% & (35/76)\\\hline
\multicolumn{12}{|c|}{Experiment using ``X {\it no} Y'' and verb case frame}\\\hline
 91\% & (60/66) & 86\% & (60/70) & 66\% & (23/35) & 79\% & (23/29) & 82\% & (83/101)& 84\% & (83/99)\\\hline
 63\% & (24/38) & 83\% & (24/29) & 63\% & (20/32) & 56\% & (20/36) & 63\% & (44/70) & 68\% & (44/65)\\\hline
\multicolumn{12}{|c|}{Estimation for the hypothetical use of a noun case frame dictionary}\\\hline
 91\% & (60/66) & 88\% & (60/68) & 69\% & (24/35) & 89\% & (24/27) & 83\% & (84/101)& 88\% & (84/95)\\\hline
 79\% & (30/38) & 86\% & (30/35) & 63\% & (20/32) & 77\% & (20/26) & 71\% & (50/70) & 82\% & (50/61)\\\hline
\end{tabular}
\end{center}


The upper row and the lower row of this table 
show rates on training sentences and 
 test sentences respectively. 

The training sentences are used to 
set the values given in the rules (Section~\ref{sec:ref_pro}) by hand. \\
{
Training sentences \{example sentences \cite{walker2} (43 sentences), a folk tale {\it Kobutori jiisan} \cite{kobu} (93 sentences), an essay in {\it Tenseijingo} (26 sentences), an editorial (26 sentences)\}\\
Test sentences \{a folk tale {\it Tsuru no ongaeshi} \cite{kobu} (91 sentences), two essays in {\it Tenseijingo} (50 sentences), an editorial (30 sentences)\}

{\it Precision}\, is 
the fraction  of the noun phrases which were judged 
to have the indirect anaphora as antecedents. 
{\it Recall}\, is the fraction of the noun phrases 
 which have the antecedents of indirect anaphora. 
We use precision and recall to evaluate 
because the system judges that 
a noun which is not an antecedent of indirect anaphora 
is an antecedent of indirect anaphora, 
and we check these errors thoroughly.}
\end{minipage}
}
\end{table*}

We show the result of anaphora resolution 
using both ``X {\it no} Y'' and a verb case frame dictionary
in Table \ref{tab:sougoukekka}. 
We obtained a recall rate of 63\% and 
a precision rate of 68\% when we estimated indirect anaphora 
in test sentences. 
This indicates that 
the information of ``X {\it no} Y'' 
is useful to a certain extent 
even though we cannot make use of a noun frame dictionary. 
We also tested 
the system when it did not have 
any semantic information. 
The precision and the recall were lower. 
This indicates that semantic information is necessary. 
The experiment was performed by 
fixing all the semantic similarity values $S$ to 0. 

We also estimated the results 
for the hypothetical 
use of a noun case frame dictionary. 
We estimated these results in the following manner: 
We looked over the errors 
that had occured when we used 
``X {\it no} Y'' and a verb case frame dictionary. 
We regarded 
errors made for one of the following three reasons as right answers: 
\begin{enumerate}
\item 
Proper examples do not exist in examples of ``X {\it no} Y'' 
or in the verb case frame dictionary.  
\item 
Wrong examples exist in examples of ``X {\it no} Y'' 
or in the verb case frame dictionary. 
\item 
A noun case frame is different from a verb case frame. 
\end{enumerate}
If we were to make a noun case frame dictionary, 
it would have some errors, and 
the success ratio would be lower than 
the ratio shown in Table \ref{tab:sougoukekka}. 

\subsection*{Discussion of Errors}

Even if we had a noun case frame dictionary, 
there are certain pairs of nouns in indirect anaphoric relationship
that could not be resolved using our framework. 
\begin{equation}
  \begin{minipage}[h]{7.5cm}
\small
{\it kon'na hidoi hubuki-no naka-wo ittai dare-ga kita-no-ka-to 
ibukarinagara, obaasan-wa iimashita.}\\
(Wondering who could have come in such a heavy snowstorm, 
the old woman said:) \\
{\it ``donata-jana''}\\
(``Who is it?'')\\
{\it to-wo aketemiruto, soko-niwa 
zenshin yuki-de masshironi natta \underline{musume}-ga 
tatte orimashita.}\\
(She opened the door, and there stood before 
her \underline{a girl} all covered with snow. )
\end{minipage}
\end{equation}
The underlined {\it musume} has 
two main meanings: a daughter or a girl. 
In the above example, 
{\it musume} means ``girl'' 
and has no indirect anaphora relation 
but the system incorrectly judged that 
it is the daughter of {\it obaasan} (the old woman). 
This is a problem of noun role ambiguity 
and is very difficult to solve. 

\begin{table*}[t]
    \caption{Examples of arranged ``X {\it no} Y''}
    \label{tab:noun_bgh}
  \begin{center}
\begin{tabular}{|p{3cm}|p{12.5cm}|}\hline
Noun Y          & Arranged noun X\\\hline
{\it kokumin} (nation)    &  $<$Human$>$  {\it aite} (partner) \, $<$Organization$>$ {\it kuni} (country), {\it senshinkoku} (an advanced country), {\it ryoukoku} (the two countries), {\it naichi} (inland), {\it zenkoku} (the whole country), {\it nihon} (Japan), {\it soren} (the Soviet Union),
 {\it eikoku} (England), {\it amerika} (America), {\it suisu} (Switzerland), {\it denmaaku} (Denmark), {\it sekai} (the world)\\\hline
{\it genshu} (the head of state)    &  $<$Human$>$ {\it raihin} (visitor) \, $<$Organization$>$ {\it gaikoku} (a foreign country), {\it kakkoku} (each country), {\it poorando} (Poland)\\\hline
{\it yane} (roof)    &  $<$Organization$>$ {\it hokkaido} (Hokkaido), {\it sekai} (the world), {\it gakkou} (school), {\it koujou} (factory), {\it gasorinsutando} (gas station), {\it suupaa} (supermarket),
 {\it jitaku} (one's home), {\it honbu} (the head office) \, $<$Product$>$ {\it kuruma} (car), {\it juutaku} (housing), {\it ie} (house), {\it shinden} (temple), {\it genkan} (entrance),
{\it  shinsha} (new car) $<$Phenomenon$>$ {\it midori} (green) $<$Action$>$ {\it kawarabuki} (tile-roofed)
 $<$Mental$>$ {\it houshiki} (method) $<$Character$>$ {\it keishiki} (form)\\\hline
{\it mokei} (model)  &  $<$Animal$>$ {\it zou} (elephant) \,$<$Nature$>$ {\it fujisan} (Mt. Fuji) \,$<$Product$>$ {\it imono} (an article of cast metal), {\it manshon} (an apartment house),
{\it  kapuseru} (capsule), {\it densha} (train), {\it hune} (ship), {\it gunkan} (warship), {\it hikouki} (airplane), {\it jettoki} (jet plane) \,$<$Action$>$ {\it zousen} (shipbuilding)
 $<$Mental$>$ {\it puran} (plan) \, $<$Character$>$ {\it unkou} (movement)\\\hline
{\it gyouji} (event)  &  $<$Human$>$ {\it koushitsu} (the Imperial Household), {\it oushitsu} (a Royal family), {\it iemoto} (the head of a school) \, $<$Organization$>$ {\it nouson} (an agricultural village), {\it ken} (prefecture), {\it nihon} (Japan),
{\it soren} (the Soviet Union), {\it tera} (temple), {\it gakkou} (school) \, $<$Action$>$ {\it shuunin} (take up one's post), {\it matsuri} (festival),
{\it  iwai} (celebration), {\it junrei} (pilgrimage) \, $<$Mental$>$ {\it kourei} (an established custom), {\it koushiki} (formal) \\\hline
{\it jinkaku} (personality) &  $<$Human$>$ {\it watashi} (myself), {\it ningen} (human), {\it seishounen} (young people), {\it seijika} (statesman) \\\hline
\end{tabular}
\end{center}
\end{table*}

The following example also presents a difficult problem: 
\begin{equation}
  \begin{minipage}[h]{15.5cm}
\small
    \begin{tabular}[t]{lll}
{\it shushou-wa} & {\it \underline{teikou}-no} & {\it { {\it tsuyoi}}} \\
(prime minister) & ({resistance}) & (very hard)\\
{\it senkyoku-no} & { {\it kaishou-wo}} & { {\it miokutta}}.\\
(electoral district) & (modification) & (give up)\\
\multicolumn{3}{p{7.5cm}}{
  (The prime minister gave up the modification of some electoral districts where \underline{the resistance} was very hard.)}
    \end{tabular}
  \end{minipage}
\label{eqn:nininku_ayamari}
\end{equation}
On the surface, the underlined {\it teikou} (resistance) appears 
to refer indirectly to {\it senkyoku} (electoral district). 
But actually 
{\it teikou} (resistance) refers to the candidates 
of {\it senkyoku} (electoral district) 
not to {\it senkyoku} (electoral district) itself. 
To arrive at this conclusion, 
in other words, to connect {\it senkyoku} (electoral district) and 
{\it teikou} (resistance), 
it is necessary to use a two-step relation, 
``an electoral district $\Rightarrow$ candidates,'' 
``candidates $\Rightarrow$ resist'' in sequence. 
It is not easy, however, to change our system 
so it can deal with two-step relationships. 
If we apply the use of two-step relationships to nouns, 
many nouns which are not in an indirect anaphoric relation 
will be incorrectly judged as indirect anaphora. 
A new method is required in order to infer two relationships in sequence.

\section{Consideration of Construction of Noun Case Frame Dictionary}

We used ``X {\it no} Y'' (Y of X) to 
resolve indirect anaphora. 
But we would achieve get a higher accuracy rate 
if we could utilize a good noun case frame dictionary. 
Therefore we have to consider 
how to construct a noun case frame dictionary. 
A key is to get the detailed meaning of ``{\it no} (of)'' in ``X {\it no} Y.'' 
If it is automatically obtainable, 
a noun case frame dictionary could be constructed automatically. 
Even if the semantic analysis of ``X {\it no} Y'' is not done well, 
we think that it is still possible 
to construct the dictionary using ``X {\it no} Y.'' 
For example, 
we arrange ``noun X {\it no} noun Y'' by 
the meaning of ``noun Y,'' 
arrange them by the meaning of ``noun X'', 
delete those where 
``noun X'' is an adjective noun, 
and obtain the results shown in Table \ref{tab:noun_bgh}. 
In this case, we use 
the thesaurus dictionary ``{\it Bunrui Goi Hyou}'' \cite{bgh} 
to learn the meanings of nouns. 
It should not be difficult to construct a noun case frame dictionary 
by hand using Table \ref{tab:noun_bgh}. 
We will make a noun case frame dictionary 
by removing {\it aite} (partner) in the line of {\it kokumin} (nation), 
{\it raihin} (visitor) in the line of {\it genshu} (the head of state), 
and noun phrases which mean characters and features. 
When we look over the noun phrases for {\it kokumin} (nation), 
we notice that almost all of them refer to countries. 
So we will also make the semantic constraint 
(or the semantic preference) 
that countries can be connected to {\it kokumin} (nation). 
When we make a noun case frame dictionary, 
we must remember that 
examples of ``X {\it no} Y'' are insufficient 
and we must add examples. 
For example, in the line of {\it genshu} (the head of state) 
there are few nouns that mean countries. 
In this case, 
it is good to add examples by from the arranged nouns for {\it kokumin} (nation), 
which is similar to {\it genshu} (the head of state). 
Since in this method 
examples are arranged by meaning in this method, 
it will not be very difficult to add examples. 

\section{Conclusion}

We presented how to resolve indirect anaphora in Japanese nouns. 
We need a noun case frame dictionary 
containing information about noun relations
to analyze indirect anaphora, 
but no such dictionary exists at present. 
Therefore, we used examples of ``X {\it no} Y'' (Y of X) 
and a verb case frame dictionary. 
We estimated indirect anaphora 
by using this information, 
and obtained a recall rate of 63\% and 
a precision rate of 68\% on  test sentences. 
This indicates that 
information about ``X {\it no} Y'' 
is useful 
when we cannot make use of a noun case frame dictionary. 
We estimated the results that would be given by 
a noun case frame dictionary, 
and obtained recall and precision rates of 
71\% and 82\% respectively. 
Finally, we proposed a way to construct a noun case frame dictionary 
by using examples of ``X {\it no} Y.'' 


\end{document}